\documentclass[]{llncs}

\usepackage[T1]{fontenc}
\usepackage{lmodern}
\usepackage{amssymb,amsmath}
\usepackage{ifxetex,ifluatex}
\usepackage{fixltx2e} 

\usepackage{makeidx}
\institute{
      DETI - Institute of Electronics and Informatics Engineering of Aveiro \\
      University of Aveiro, Portugal \\
   \texttt{\{eduardopinho, carlos.costa\}@ua.pt}}

\providecommand{\tightlist}{%
  \setlength{\itemsep}{0pt}\setlength{\parskip}{0pt}}

\IfFileExists{upquote.sty}{\usepackage{upquote}}{}
\ifnum 0\ifxetex 1\fi\ifluatex 1\fi=0 
  \usepackage[utf8]{inputenc}
\else 
  \ifxetex
    \usepackage{mathspec}
    \usepackage{xltxtra,xunicode}
  \else
    \usepackage{fontspec}
  \fi
  \defaultfontfeatures{Mapping=tex-text,Scale=MatchLowercase}
  
\fi
\IfFileExists{microtype.sty}{\usepackage{microtype}}{}
\usepackage{longtable,booktabs}
\usepackage{graphicx}
\makeatletter
\def\ScaleIfNeeded{%
  \ifdim\Gin@nat@width>\linewidth
    \linewidth
  \else
    \Gin@nat@width
  \fi
}
\makeatother
\let\Oldincludegraphics\includegraphics
{%
 \catcode`\@=11\relax%
 \gdef\includegraphics{\@ifnextchar[{\Oldincludegraphics}{\Oldincludegraphics[width=\ScaleIfNeeded]}}%
}%
\ifxetex
  \usepackage[setpagesize=false, 
              unicode=false, 
              xetex]{hyperref}
\else
  \usepackage[unicode=true]{hyperref}
\fi
\hypersetup{breaklinks=true,
            bookmarks=true,
            pdfauthor={Eduardo Pinho; Carlos Costa},
            pdftitle={Unsupervised learning for concept detection in medical images: a comparative analysis},
            colorlinks=true,
            citecolor=blue,
            urlcolor=blue,
            linkcolor=magenta,
            pdfborder={0 0 0}}
\title{Unsupervised learning for concept detection in medical images: a
comparative analysis}
\author{Eduardo Pinho \and Carlos Costa}
\date{}
\usepackage{graphicx}
\usepackage{subfig}
\usepackage{caption}
\makeatletter
\@ifpackageloaded{subfig}{}{\usepackage{subfig}}
\@ifpackageloaded{caption}{}{\usepackage{caption}}
\AtBeginDocument{%

}
\AtBeginDocument{%

}
\@ifpackageloaded{float}{}{\usepackage{float}}
\floatstyle{ruled}
\@ifundefined{c@chapter}{\newfloat{codelisting}{h}{lop}}{\newfloat{codelisting}{h}{lop}[chapter]}
\floatname{codelisting}{Listing}

\makeatother

\begin{document}
\maketitle
\begin{abstract}
As digital medical imaging becomes more prevalent and archives increase
in size, representation learning exposes an interesting opportunity for
enhanced medical decision support systems. On the other hand, medical
imaging data is often scarce and short on annotations. In this paper, we
present an assessment of unsupervised feature learning approaches for
images in the biomedical literature, which can be applied to automatic
biomedical concept detection. Six unsupervised representation learning
methods were built, including traditional bags of visual words,
autoencoders, and generative adversarial networks. Each model was
trained, and their respective feature space evaluated using images from
the ImageCLEF 2017 concept detection task. We conclude that it is
possible to obtain more powerful representations with modern deep
learning approaches, in contrast with previously popular computer vision
methods. Although generative adversarial networks can provide good
results, they are harder to succeed in highly varied data sets. The
possibility of semi-supervised learning, as well as their use in medical
information retrieval problems, are the next steps to be strongly
considered.


\end{abstract}

\hypertarget{introduction}{%
\section{Introduction}\label{introduction}}

In an era of a steadly increasing use of digital medical imaging, image
recognition poses an interesting prospect for novel solutions supporting
clinicians and researchers. In particular, the representation learning
field is growing fast in recent years {[}1{]}, and many of the
breakthroughs in this field are occurring in deep learning methods,
which have also been strongly considered in healthcare {[}2{]}.
Leveraging representation learning tools to the medical imaging field is
feasible and worthwhile, as they can provide additional levels of
introspection of clinical cases through content-based image retrieval
(CBIR).

Multiple initiatives for the provision of medical imaging data sets
exist, the process of annotating the data with useful information is
exhaustive and requires medical expertise, as it often nails down to a
medical diagnosis. In the face of few to no annotations, unsupervised
learning stands as a possible means of feature extraction for a
measurement of relevance, leading to more powerful information retrieval
and decision support solutions in digital medical imaging.

Although unsupervised representation is limited for specific
classification tasks when compared to supervised learning approaches,
the latter requires an exhaustive process from experts to obtain
annotated content. Unsupervised learning, which avoids this issue, can
also provide a few other benefits, including transferrability to other
problems or domains, and can often be bridged to supervised and
semi-supervised techniques. We have hypothesized that a sufficiently
powerful representation of images would enable a medical imaging archive
to automatically detect biomedical concepts with some level of certainty
and efficiency, thus improving the system's information retrieval
capabilities over non-annotated data.

In this work, we present an assessment of unsupervised mid-level
representation learning approaches for images in the biomedical
literature. Representations are built using an ensemble of images from
biomedical literature. The learned representations were validated with a
brief qualitative feature analysis, and by training simple classifiers
for the purpose of biomedical concept detection. We show that feature
learning techniques based on deep neural networks can outperform
techniques that were previously common-place in image recognition, and
that models with adversarial networks, albeit harder to train, can
improve the quality of feature learning.

\hypertarget{related-work}{%
\section{Related Work}\label{related-work}}

\emph{Representation learning}, or \emph{feature learning}, can be
defined as the process of learning a transformation from a data domain
into a representation that makes other machine learning tasks easier to
approach {[}1{]}. The concept of \emph{feature extraction} can be
employed when this mapping is obtained with handcrafted algorithms
rather than learned from the original data in the distribution.
Representation learning can be achieved using a wide range of methods,
such as k-means clustering, sparse coding {[}3{]} and Restricted
Boltzmann Machines (RBMs) {[}4{]}. In image recognition, algorithms
based on bags of visual words have been prevalent, as they have shown
superior results over other low-level visual feature extraction
techniques {[}5{]}, {[}6{]}. More recently however, image recognition
has had a strong focus on deep learning techniques, often with
impressive results. Among these, approaches based on autoencoders
{[}7{]}, {[}8{]} have been considered and are still prevalent to this
day.

Research on representation learning is even more intense on the
ground-breaking concept of generative adversarial networks (GANs)
{[}9{]}. GANs devise a min-max game where a \emph{generator} of ``fake''
data samples attempts to fool a \emph{discriminator} network, which in
turn learns to discriminate fake samples from real ones. As the two
components mutually improve, the generator will ultimately produce
visually-appealing samples that are similar to the original data. The
impressive quality of the samples generated by GANs have led the
scientific community into devising new GAN variants and applications to
this adversarial loss, including for feature learning {[}10{]}.

Representation learning has been notably used in medical image
retrieval, although even in this decade, handcrafted visual feature
extraction algorithms are frequently considered in this context
{[}11{]}, {[}12{]}. Nonetheless, although the interest in deep learning
is relatively recent, a wide variety of neural networks have been
studied for medical image analysis {[}13{]}, as they often exhibit
greater potential for the task {[}14{]}. The use of unsupervised
learning techniques is also well regarded as a means of exploiting as
much of the available medical imaging data as possible {[}15{]}. On the
other hand, the amount of medical imaging data may be scarse for many
use cases, which makes training deep neural networks a difficult
process.

\hypertarget{methods}{%
\section{Methods}\label{methods}}

We have considered a set of unsupervised representation learning
techniques, both traditional (as in, employing classic computer vision
algorithms) and based on deep learning, for the scope of images in the
biomedical domain. These representations were subsequently used for the
task of biomedical concept detection. Namely:

\begin{itemize}
\tightlist
\item
  We have experimented with creating image descriptors using bags of
  visual words (BoWs), for two different visual keypoint extraction
  algorithms.
\item
  With the use of modern deep learning approaches, we have designed and
  trained various deep neural network architectures: a sparse denoising
  autoencoder, (SDAE), a variational autoencoder (VAE), a bidirectional
  generative adversarial network (BiGAN), and an adversarial autoencoder
  (AAE).
\end{itemize}

\hypertarget{sec:bows}{%
\subsection{Bags of Visual Words}\label{sec:bows}}

For each data set, images were converted to greyscale without resizing
and visual keypoint descriptors were subsequently extracted. We employed
two keypoint extraction algorithms separately: Scale Invariant Feature
Transform (SIFT) {[}16{]}, and Oriented FAST and Rotated BRIEF (ORB)
{[}17{]}. While both algorithms obtain scale and rotation invariant
descriptors, ORB is known to be faster and require less computational
resources. The keypoints were extracted and their respective descriptors
computed using OpenCV {[}18{]}. Each image would yield a variable number
of descriptors of fixed size (128-dimensional for SIFT, 32-dimensional
for ORB). In cases where the algorithm did not retrieve any keypoints,
the algorithm's parameters were adjusted to loosen edge detection
criteria. All procedures described henceforth are the same for both ORB
and SIFT keypoint descriptors.

From the training set, 3000 files were randomly chosen and their
respective keypoint descriptors collected to serve as template
keypoints. A visual vocabulary (codebook) of size \(k = 512\) was then
obtained by performing k-means clustering on all template keypoint
descriptors and retrieving the centroids of each cluster, yielding a
list of 512 keypoint descriptors \(\mathcal{V} = \{ V_i \}\).

Once a visual vocabulary was available, we constructed an image's BoW by
determining the closest visual vocabulary point and incrementing the
corresponding position in the BoW for each image keypoint descriptor. In
other words, for an image's BoW \(B = \{ o_i \}\), for each image
keypoint descriptor \(d_j\), \(o_i\) is incremented when the smallest
Euclidean distance from \(d_j\) to all other visual vocabulary points in
\(V\) is the distance to \(V_i\). Finally, each BoW was normalized so
that all elements lie between 0 and 1. We can picture the bag of visual
words as a histogram of visual descriptor occurrences, which can be used
as a global image descriptor {[}19{]}.

\hypertarget{deep-representation-learning}{%
\subsection{Deep Representation
Learning}\label{deep-representation-learning}}

Modern representation techniques often rely on deep learning methods. We
have considered a set of deep convolutional neural network architectures
for inferring a late feature space over biomedical images. These models
are composed of parts with very similar numbers of layers and
parameters, in order to obtain a fairer comparison in the evaluation
phase. This also means that the models will have very similar prediction
times.

Training samples were obtained through the following process: images
were resized so that its shorter dimension (width or height) was exactly
\(s_g\) pixels. Afterwards, the sample was augmented by feeding the
networks random crops of size \(s \times s\) (out of 9 possible kinds of
crops: 4 corners, 4 edges and center). Validation images were resized to
fit the \(s \times s\) dimensions. For all cases, the images' pixel RGB
values were normalized to fit in the range {[}-1, 1{]}. Unless otherwise
mentioned, the networks assumed a rescale size to \(s_g = 96\) and a
crop size \(s = 64\).

Models with an enconding or discrimination process for visual data were
based on the same convolutional neural network architecture, described
in Table~\ref{tbl:enc} and Table~\ref{tbl:dec}. These models were
influenced by the work in deep convolutional generative adversarial
networks {[}20{]}. Each encoder layer is composed of a 2D convolution,
followed by an optional (case-dependent) normalization algorithm and a
model-dependent non-linearity. At the top of the network, global average
pooling is performed, followed by a fully connected layer, yielding the
code tensor \(z\). The Details column in both tables may include the
normalization and activation layers that follow a convolution layer.

\hypertarget{tbl:enc}{}
\begin{longtable}[]{@{}cccl@{}}
\caption{\label{tbl:enc}A tabular representation of the SimpleNet
layers' specifications. The Details column may include the normalization
and activation layers that follow a convolution layer (where LN stands
for layer normalization and ReLU is the rectified linear unit
\(max(0, x)\)).}\tabularnewline
\toprule
\begin{minipage}[b]{0.12\columnwidth}\centering
Layer\strut
\end{minipage} & \begin{minipage}[b]{0.12\columnwidth}\centering
Kernels\strut
\end{minipage} & \begin{minipage}[b]{0.17\columnwidth}\centering
Size/Stride\strut
\end{minipage} & \begin{minipage}[b]{0.39\columnwidth}\raggedright
Details\strut
\end{minipage}\tabularnewline
\midrule
\endfirsthead
\toprule
\begin{minipage}[b]{0.12\columnwidth}\centering
Layer\strut
\end{minipage} & \begin{minipage}[b]{0.12\columnwidth}\centering
Kernels\strut
\end{minipage} & \begin{minipage}[b]{0.17\columnwidth}\centering
Size/Stride\strut
\end{minipage} & \begin{minipage}[b]{0.39\columnwidth}\raggedright
Details\strut
\end{minipage}\tabularnewline
\midrule
\endhead
\begin{minipage}[t]{0.12\columnwidth}\centering
conv1\strut
\end{minipage} & \begin{minipage}[t]{0.12\columnwidth}\centering
64\strut
\end{minipage} & \begin{minipage}[t]{0.17\columnwidth}\centering
5x5 /2\strut
\end{minipage} & \begin{minipage}[t]{0.39\columnwidth}\raggedright
Normalization + Non-Linearity\strut
\end{minipage}\tabularnewline
\begin{minipage}[t]{0.12\columnwidth}\centering
conv2\strut
\end{minipage} & \begin{minipage}[t]{0.12\columnwidth}\centering
128\strut
\end{minipage} & \begin{minipage}[t]{0.17\columnwidth}\centering
5x5 /2\strut
\end{minipage} & \begin{minipage}[t]{0.39\columnwidth}\raggedright
Normalization + Non-Linearity\strut
\end{minipage}\tabularnewline
\begin{minipage}[t]{0.12\columnwidth}\centering
conv3\strut
\end{minipage} & \begin{minipage}[t]{0.12\columnwidth}\centering
256\strut
\end{minipage} & \begin{minipage}[t]{0.17\columnwidth}\centering
5x5 /2\strut
\end{minipage} & \begin{minipage}[t]{0.39\columnwidth}\raggedright
Normalization + Non-Linearity\strut
\end{minipage}\tabularnewline
\begin{minipage}[t]{0.12\columnwidth}\centering
conv4\strut
\end{minipage} & \begin{minipage}[t]{0.12\columnwidth}\centering
512\strut
\end{minipage} & \begin{minipage}[t]{0.17\columnwidth}\centering
5x5 /2\strut
\end{minipage} & \begin{minipage}[t]{0.39\columnwidth}\raggedright
Normalization + Non-Linearity\strut
\end{minipage}\tabularnewline
\begin{minipage}[t]{0.12\columnwidth}\centering
conv5\strut
\end{minipage} & \begin{minipage}[t]{0.12\columnwidth}\centering
512\strut
\end{minipage} & \begin{minipage}[t]{0.17\columnwidth}\centering
5x5 /2\strut
\end{minipage} & \begin{minipage}[t]{0.39\columnwidth}\raggedright
Normalization + Non-Linearity\strut
\end{minipage}\tabularnewline
\begin{minipage}[t]{0.12\columnwidth}\centering
avgpool\strut
\end{minipage} & \begin{minipage}[t]{0.12\columnwidth}\centering
N/A\strut
\end{minipage} & \begin{minipage}[t]{0.17\columnwidth}\centering
N/A\strut
\end{minipage} & \begin{minipage}[t]{0.39\columnwidth}\raggedright
\strut
\end{minipage}\tabularnewline
\begin{minipage}[t]{0.12\columnwidth}\centering
fc\strut
\end{minipage} & \begin{minipage}[t]{0.12\columnwidth}\centering
\(nb\)\strut
\end{minipage} & \begin{minipage}[t]{0.17\columnwidth}\centering
\strut
\end{minipage} & \begin{minipage}[t]{0.39\columnwidth}\raggedright
Linear activation\strut
\end{minipage}\tabularnewline
\bottomrule
\end{longtable}

The decoding blocks replicate the encoding process in inverse order
(Table~\ref{tbl:dec}). It starts with a fully connected network from the
latent (or prior) code vector into a series of convolutional layers.
Convolutions in these blocks are transposed (also called
\emph{fractionally-strided convolution} in literature, and
\emph{deconvolution} in a few other papers). The weights of the network
were randomly initialized with a Gaussian distribution.

\hypertarget{tbl:dec}{}
\begin{longtable}[]{@{}cccc@{}}
\caption{\label{tbl:dec}A tabular representation of the
decoder/generator layers' specifications. The Details column may include
the normalization and activation layers that follow a convolution
layer.}\tabularnewline
\toprule
\begin{minipage}[b]{0.12\columnwidth}\centering
Layer\strut
\end{minipage} & \begin{minipage}[b]{0.13\columnwidth}\centering
Kernels\strut
\end{minipage} & \begin{minipage}[b]{0.14\columnwidth}\centering
Size /stride\strut
\end{minipage} & \begin{minipage}[b]{0.49\columnwidth}\centering
Details\strut
\end{minipage}\tabularnewline
\midrule
\endfirsthead
\toprule
\begin{minipage}[b]{0.12\columnwidth}\centering
Layer\strut
\end{minipage} & \begin{minipage}[b]{0.13\columnwidth}\centering
Kernels\strut
\end{minipage} & \begin{minipage}[b]{0.14\columnwidth}\centering
Size /stride\strut
\end{minipage} & \begin{minipage}[b]{0.49\columnwidth}\centering
Details\strut
\end{minipage}\tabularnewline
\midrule
\endhead
\begin{minipage}[t]{0.12\columnwidth}\centering
fc\strut
\end{minipage} & \begin{minipage}[t]{0.13\columnwidth}\centering
4096\strut
\end{minipage} & \begin{minipage}[t]{0.14\columnwidth}\centering
\strut
\end{minipage} & \begin{minipage}[t]{0.49\columnwidth}\centering
Reshaped to 1024x2x2\strut
\end{minipage}\tabularnewline
\begin{minipage}[t]{0.12\columnwidth}\centering
dconv5\strut
\end{minipage} & \begin{minipage}[t]{0.13\columnwidth}\centering
512\strut
\end{minipage} & \begin{minipage}[t]{0.14\columnwidth}\centering
5x5 /2\strut
\end{minipage} & \begin{minipage}[t]{0.49\columnwidth}\centering
Normalization + ReLU\strut
\end{minipage}\tabularnewline
\begin{minipage}[t]{0.12\columnwidth}\centering
dconv4\strut
\end{minipage} & \begin{minipage}[t]{0.13\columnwidth}\centering
256\strut
\end{minipage} & \begin{minipage}[t]{0.14\columnwidth}\centering
5x5 /2\strut
\end{minipage} & \begin{minipage}[t]{0.49\columnwidth}\centering
Normalization + ReLU\strut
\end{minipage}\tabularnewline
\begin{minipage}[t]{0.12\columnwidth}\centering
dconv3\strut
\end{minipage} & \begin{minipage}[t]{0.13\columnwidth}\centering
128\strut
\end{minipage} & \begin{minipage}[t]{0.14\columnwidth}\centering
5x5 /2\strut
\end{minipage} & \begin{minipage}[t]{0.49\columnwidth}\centering
Normalization + ReLU\strut
\end{minipage}\tabularnewline
\begin{minipage}[t]{0.12\columnwidth}\centering
dconv2\strut
\end{minipage} & \begin{minipage}[t]{0.13\columnwidth}\centering
64\strut
\end{minipage} & \begin{minipage}[t]{0.14\columnwidth}\centering
5x5 /2\strut
\end{minipage} & \begin{minipage}[t]{0.49\columnwidth}\centering
Normalization + ReLU\strut
\end{minipage}\tabularnewline
\begin{minipage}[t]{0.12\columnwidth}\centering
dconv1\strut
\end{minipage} & \begin{minipage}[t]{0.13\columnwidth}\centering
3\strut
\end{minipage} & \begin{minipage}[t]{0.14\columnwidth}\centering
5x5 /2\strut
\end{minipage} & \begin{minipage}[t]{0.49\columnwidth}\centering
Linear activation\strut
\end{minipage}\tabularnewline
\bottomrule
\end{longtable}

\hypertarget{sec:ae}{%
\subsubsection{Sparse Denoising Autoencoder}\label{sec:ae}}

\begin{figure}
\hypertarget{fig:sdae}{%
\centering
\includegraphics[width=9cm,height=3.8cm]{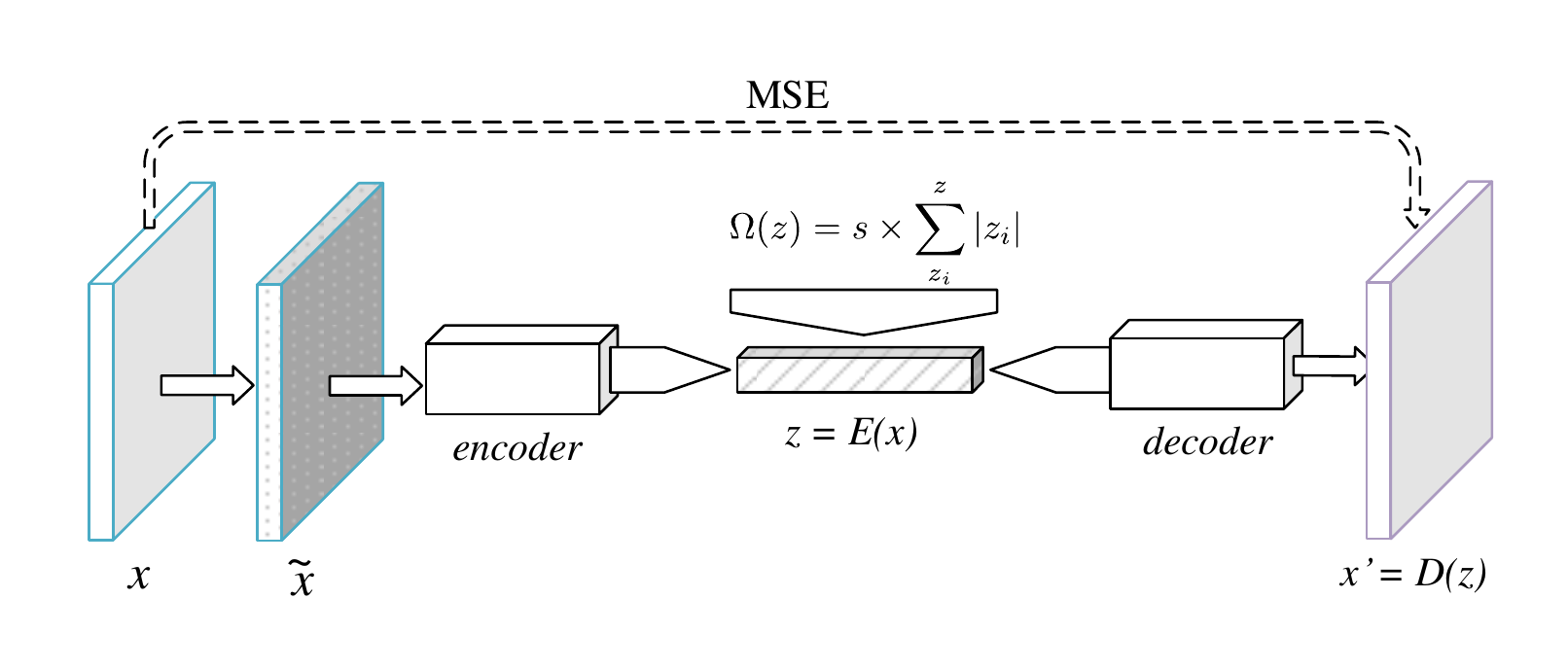}
\caption{Diagram of the sparse denoising autoencoder.}\label{fig:sdae}
}
\end{figure}

The first tested deep neural network model is a common autoencoder with
denoising and sparsity constraints (Figure~\ref{fig:sdae}). In the
training phase, a Gaussian noise of standard deviation 0.05 was applied
over the input, yielding a noisy sample \(\tilde{x}\). As a denoising
autoencoder, its goal is to learn the pair of functions \((E, D)\) so
that \(x' = D(E(\tilde{x}))\) is closest to the original input \(x\).
The aim of making \(E\) a function of \(\tilde{x}\) is to force the
process to be more stable and robust, thus leading to higher quality
representations {[}7{]}.

Sparsity was achieved with two mechanisms. First, a rectified linear
unit (ReLU) activation was used after the last fully connected layer of
the encoder, turning negative outputs from the previous layer into
zeros. Second, an absolute value penalization was applied to \(z\), thus
adding the extra minimization goal of keeping the code sum small in
magnitude. The final decoder loss function was therefore:
\[\mathcal{L}(E, D) = \frac{1}{r} \sum^{\mathrm{r}}_{i=0}{\left(x_i - x_i'\right)}^2 + \Omega(z)\]
where \[\Omega(z) = s \times \sum^{z}_{z_i}{z_i}\] is the sparsity
penalty function, \(\mathrm{r} = 64 \times 64\) is the number of pixels
in the input images, and \(x\) represents the original input without
synthesized noise. \(s\) is the sparsity coefficient, which we left
defined as \(s = 0.0001\). This network used batch normalization
{[}21{]} and (non-leaking) ReLU activations.

\hypertarget{variational-autoencoder}{%
\subsubsection{Variational Autoencoder}\label{variational-autoencoder}}

\begin{figure}
\hypertarget{fig:vae}{%
\centering
\includegraphics[width=9cm,height=3.8cm]{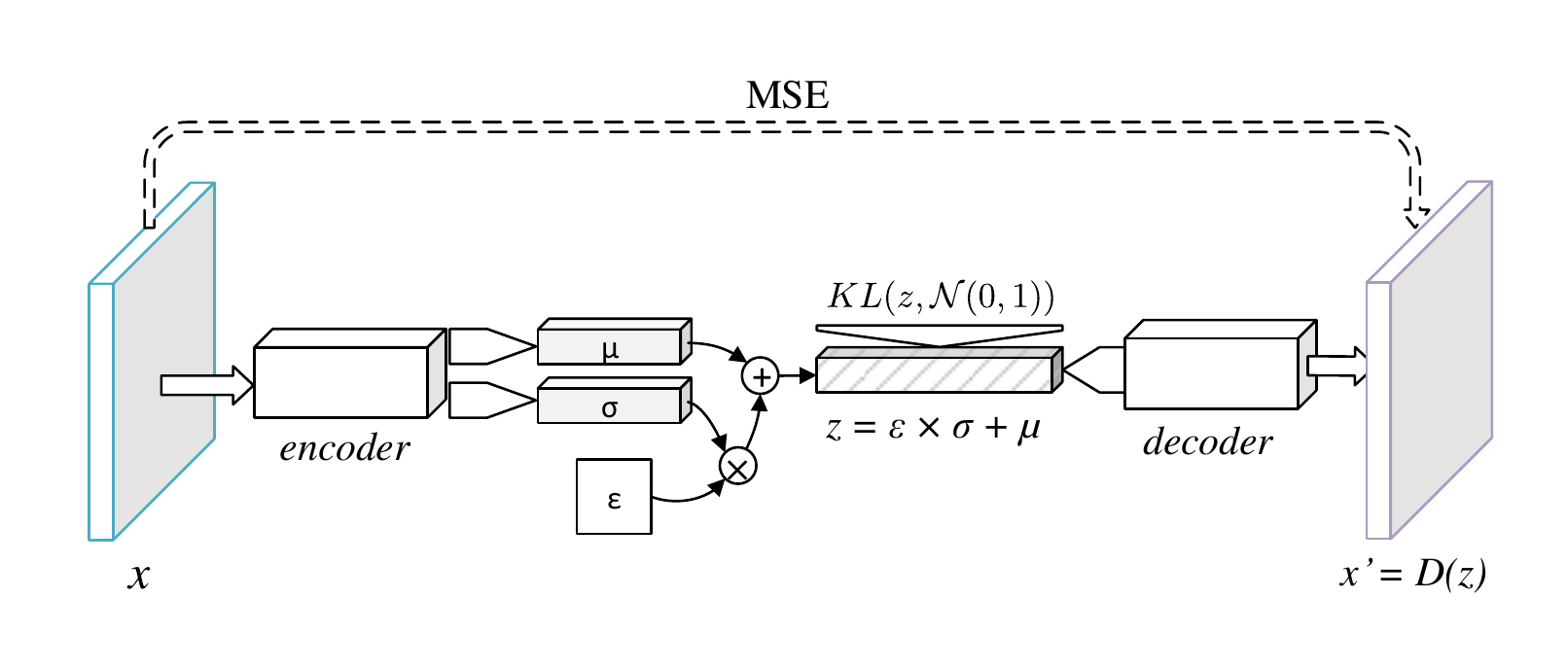}
\caption{Diagram of the variational autoencoder.}\label{fig:vae}
}
\end{figure}

The encoder of the variational autoencoder (Figure~\ref{fig:vae}) learns
a stochastic distribution which can be sampled from, by minimizing the
Kulback-Leibler divergence with a unitary normal distribution {[}8{]}.
Like in the SDAE, convolutions were followed by batch normalization
{[}21{]} and (non-leaking) ReLU activations.

\hypertarget{bidirectional-gan}{%
\subsubsection{Bidirectional GAN}\label{bidirectional-gan}}

\begin{figure}
\hypertarget{fig:bigan}{%
\centering
\includegraphics[width=9cm,height=5cm]{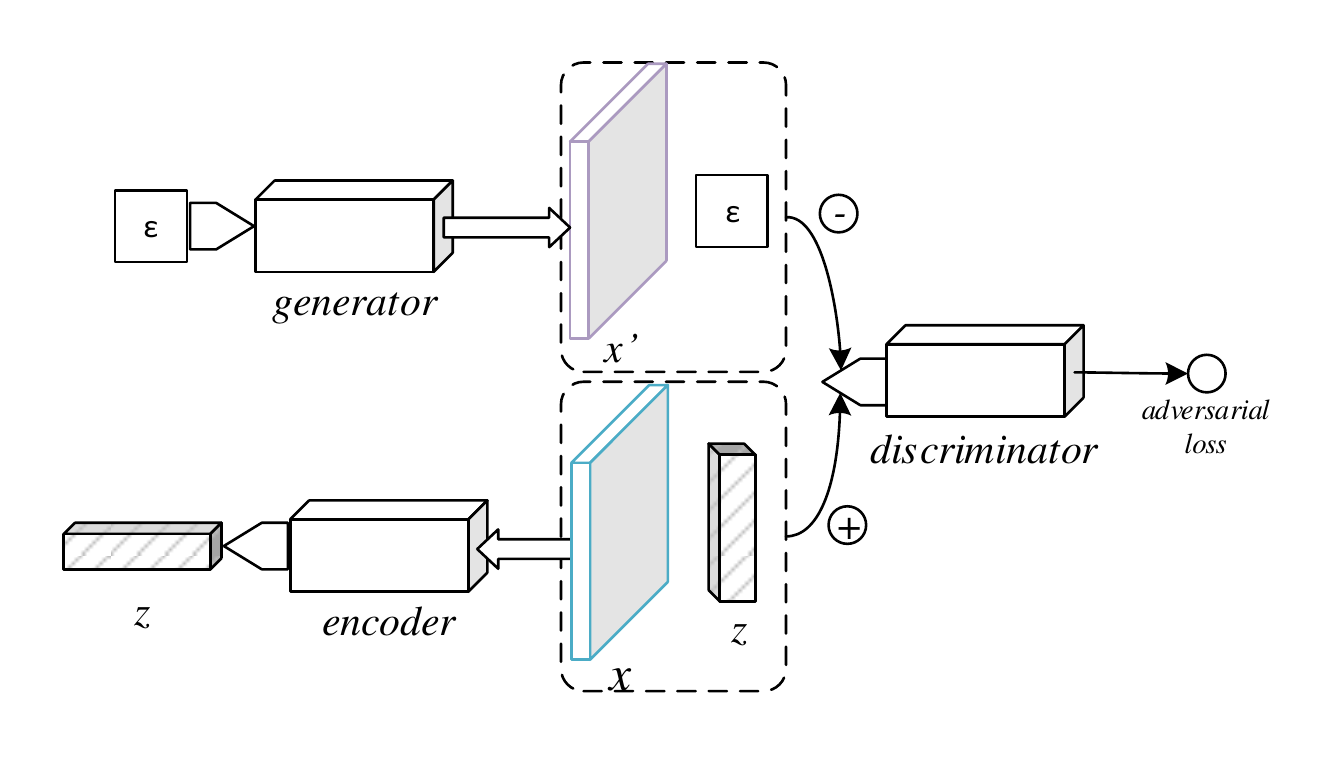}
\caption{Diagram of the bidirectional GAN.}\label{fig:bigan}
}
\end{figure}

While GANs are known to show great potential for representation
purposes, the basic GAN archicture does not provide a means to encode
samples to their respective prior. The bidirectional GAN, depicted in
Figure~\ref{fig:bigan}, addresses this concern by including an encoder
component, which learns the inverse process of the generator {[}10{]}.
Rather than only observing data samples, the BiGAN discriminator's loss
function depends on the code-sample pair.

The \emph{Encoder} component of the network used the same design as the
discriminator, with the exception that the original data was fed with a
size \(s\) of 112, the outcome of cropping the data after the shortest
dimension was resized to 128 pixels (as in, \(s_g = 128\) for the
encoder). Images were still downsampled to 64x64 to be fed to the
discriminator. Like in {[}10{]}, all constituent parts of the GAN were
optimized simultaneously in each iteration. The encoder and the
discriminator of this model used layer normalization {[}22{]} and leaky
ReLU with a leaking factor of 0.2 on all except the last respective
convolutional layers.

\hypertarget{adversarial-autoencoder}{%
\subsubsection{Adversarial Autoencoder}\label{adversarial-autoencoder}}

\begin{figure}
\hypertarget{fig:aae}{%
\centering
\includegraphics[width=8.8cm,height=4.6cm]{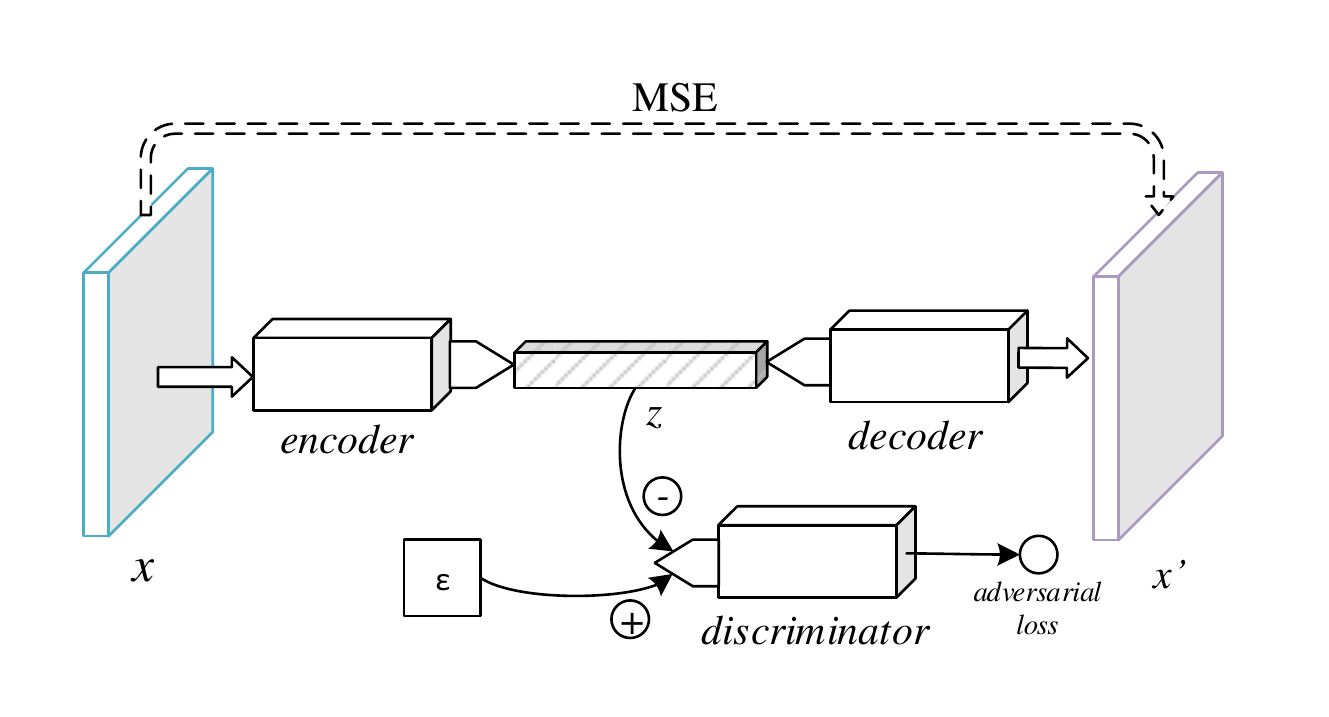}
\caption{Diagram of the adversarial autoencoder.}\label{fig:aae}
}
\end{figure}

The adversarial autoencoder (AAE) is an autoencoder in which a
discriminator is added to the bottleneck vector {[}23{]}. While reducing
the \(L_2\)-norm distance between a sample and its decoded form, the
full network includes an adversarial loss for distinguishing the
encoder's output from a stochastic prior code, thus serving as a
regularizer to the encoding process.

Our AAE used a simple code discriminator composed of 2 fully connected
layers of 128 units with a leaky ReLU activation for the first two
layers, followed by a single neuron without a non-linearity. During
training, the discriminator is fed a prior \(z\) sampled from a random
normal distribution \(\mathcal{N}(0, 1)\) as the \emph{real} code, and
the output of the encoder \(E(x)\) as the \emph{fake} code. The model
uses layer normalization {[}22{]} on all except the last layers of each
component, and leaky ReLU with a leaking factor of 0.2. Like in
{[}10{]}, all three components' parameters were updated simultaneously
in each iteration.

\hypertarget{network-training-details}{%
\subsubsection{Network Training
Details}\label{network-training-details}}

The networks were trained through stochastic gradient descent, using the
Adam optimizer {[}24{]}. The \(\alpha_1\) hyperparameter was set to 0.5
for the BiGAN and the AAE, and 0.9 for the remaining networks.

Each neural network model was trained over 206000 steps, which is
approximately 100 epochs, with a mini-batch size of 64. The base
learning rate was 0.0005. The learning rate was multiplied by 0.2
halfway through the training process (50 epochs), to facilitate
convergence.

All neural network training and latent code extraction was conducted
using TensorFlow, and TensorBoard was used during the development for
monitoring and visualization {[}25{]}. Depending on the particular
model, training took on average 120 hours (a maximum of 215 hours, for
the adversarial autoencoder) to complete on one of the GPUs of an NVIDIA
Tesla K80 graphics card in an Ubuntu server machine.

\hypertarget{evaluation}{%
\subsection{Evaluation}\label{evaluation}}

The previously described methods for representation learning were aimed
towards addressing the domain of biomedical images. A proper validation
of these features was made with the use of the data sets from the
ImageCLEF 2017 concept detection challenge {[}26{]}. As one of the
sub-tasks of the caption prediction challenge, the goal of the challenge
is to conceive a computer model for identifying the individual
components from medical images, from which full captions could be
composed. This task was accompanied with three data sets containing
various images from biomedical journals: the \emph{training set} (164614
images), the \emph{validation set} (10000 images) and the \emph{testing
set} (10000 images). These sets were annotated with the lists of
biomedical term identifiers from the UMLS (Unified Medical Language
System)\footnote{\url{https://www.nlm.nih.gov/research/umls}} vocabulary
for each image. The testing set's annotations were hidden during the
challenge, but were later on provided to participants.

Each of the set of features, learned from the approaches described in
the previous section, were used to train simple classifiers for concept
detection. In both cases, the same training and validation folds from
the original data set were considered, after being mapped to their
respective feature spaces. In addition, data points in the validation
set with an empty list of concepts were discarded.

These simple models were used to predict the concept list of each image
by sole observation of their respective feature set. Therefore, the
assessment of our representation learning methods is made based on the
effectiveness of capturing high-level features from latent codes alone.

\hypertarget{sec:classification}{%
\subsubsection{Logistic Regression}\label{sec:classification}}

Aiming for low complexity and classification speed, we performed
logistic regression with stochastic gradient descent for concept
detection, treating the UMLS terms as labels. More specifically, linear
classifiers were trained over the features, one for each of the 750
(seven-hundred and fifty) most frequently occurring concepts in the
training set. All models were trained using FTRL-Proximal optimization
{[}27{]} with a base learning rate of \(0.05\), an \(L_1\)-norm
regularization factor of \(0.001\), and a batch size of 128. Since the
biomedical concepts are very sparse and imbalanced, the \(F_1\) score
was considered as the main evaluation metric, which was calculated with
respect to multiple fixed operating point thresholds (namely, \(0.025\),
\(0.05\), \(0.075\), \(0.1\), \(0.125\), \(0.15\), \(0.175\), and
\(0.2\)) for each sample and averaged across the 750 labels. The
threshold which resulted in the highest mean \(F_1\) score on the
validation set is recorded, and the respective precision, recall, and
area under the ROC curve were also included. Subsequently, the same
model and threshold were used for predicting the concepts in the testing
set, the \(F_1\) score of which was retrieved with the official
evaluation tool from the ImageCLEF challenge.

Since it is also possible to combine multiple representations with
simple vector concatenation, we have experimented training these
classifiers using a mixture of features from the SDAE and AAE latent
codes. This process is often called \emph{early fusion}, and is
contrasted with \emph{late fusion}, which involves merging the results
of separate models. Each model undertook a few dozens of training epochs
until the best \(F_1\) score among the thresholds would no longer
improve. In practice, training and evaluation of the linear classifiers
was done with TensorFlow.

\hypertarget{sec:knn}{%
\subsubsection{\texorpdfstring{\(k\)-nearest
neighbors}{k-nearest neighbors}}\label{sec:knn}}

A relevant focus of interest in representation learning is its potential
in information retrieval. While concept detection is not a retrieval
problem, and the use of retrieval techniques is a naive approach to
classification, it is fast and scales better in the face of multiple
classes. Furthermore, it enables a rough assessment of whether the
representation would fare well in retrieval tasks where similarity
metric were not previously learned, which is the case for the Euclidean
distance between features.

A modified form of the k-nearest neighbors algorithm was used as a
second means of evaluation. Each data point in the validation set had
its concepts predicted by retrieving the \(n\) closest points from the
training feature set (henceforth called neighbors) in Euclidean space
and accumulating all concepts of those neighbors into a boolean sum of
labels. This tweak makes the algorithm more sensitive to very sparse
classification labels, such as those found in the biomedical concept
detection task. All natural numbers from 1 to 5 were tested for the
possible \(k\) number of neighbors to consider. Analogous to the
logistic regression above, the \(k\) which resulted in the highest
\(F_1\) score on the validation set was regarded as the optimal
parameter, and predictions over the testing set were evaluated using the
optimal \(k\).\\
The actual search for the nearest neighbors was performed using the
Faiss library, which contributed to a rapid retrieval {[}28{]}. Feature
fusion was not considered in the results, as they did not seem to bring
any improvement over singular representations.

\hypertarget{results}{%
\section{Results}\label{results}}

\hypertarget{qualitative-results}{%
\subsection{Qualitative results}\label{qualitative-results}}

\begin{figure}
\hypertarget{fig:pca}{%
\centering
\includegraphics{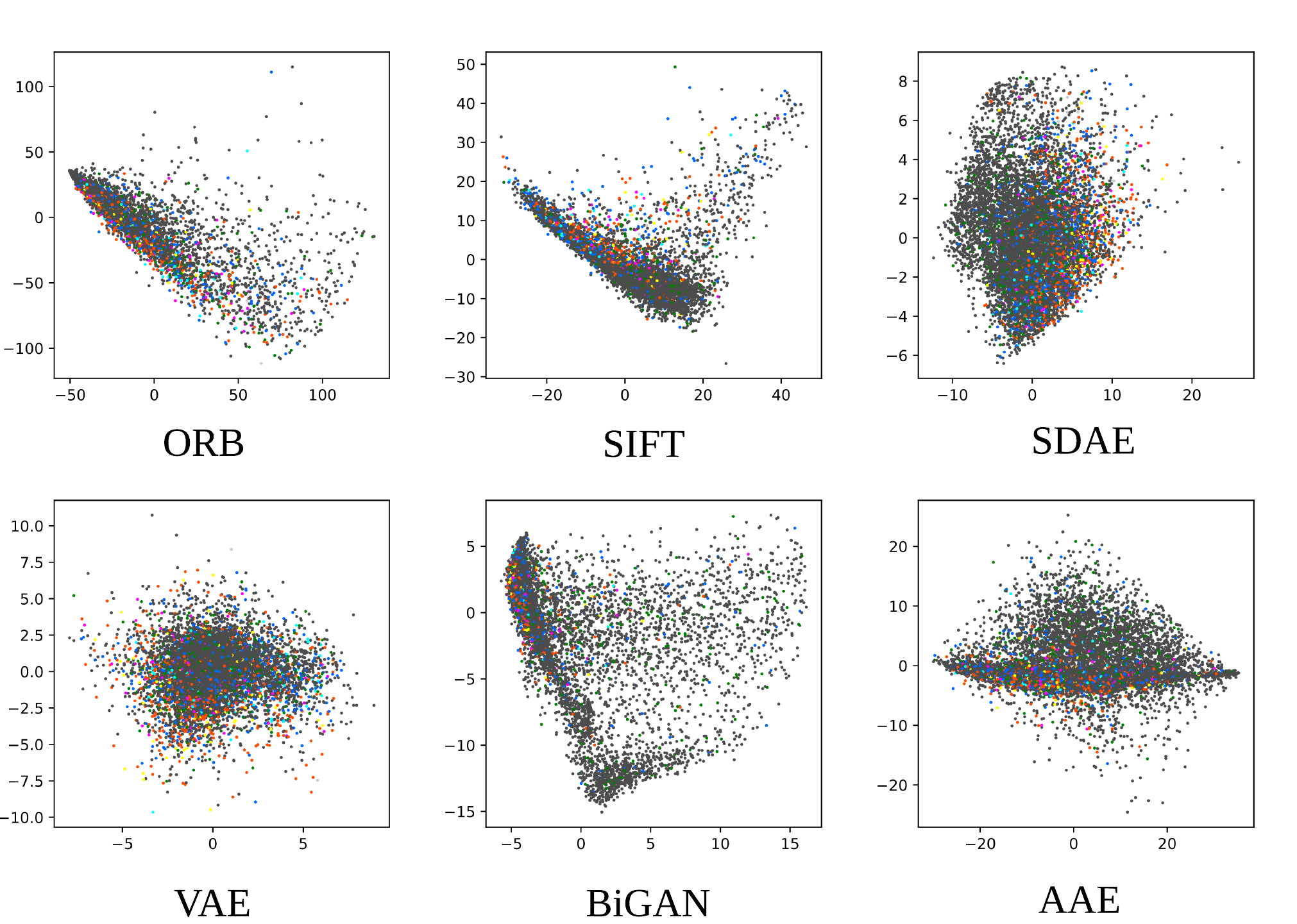}
\caption{The 2D projections of the latent codes in the validation set,
for each learned feature space. Best seen in color.}\label{fig:pca}
}
\end{figure}

Each representation learning approach described in this work resulted in
a 512-dimensional feature space. Figure~\ref{fig:pca} shows the result
of mapping the validation feature set of each representation learned
into a two-dimensional space, using principal component analysis (PCA).
The three primary colors were used (red, green, and blue) to label the
points with the three most commonly occurring UMLS terms in the training
set, namely \texttt{C1696103} (\emph{Image-dosage form}),
\texttt{C0040405} (\emph{X-Ray Computed Tomography}), and
\texttt{C0221198} (\emph{Lesion}), each painted in an additive fashion.

While extreme outliers were removed from the figures, it can be noted
that the ORB, SIFT and BiGAN representations had more outliers than the
other three representations. A good representation would enable samples
to be linearly separable based on their list of concepts. Even though
the concept detection task is too hard for a clear cut separation, one
can still identify regions in the manifold in which points of one of the
frequent labels are mostly gathered. The existence of concentrations of
random points in certain parts of the manifold, as further observed from
the classification results, is noticeable mostly in poorer quality
representations.

The latent space regularization in representations based on deep
learning is also apparent in these plots: both the AAE (with the
approximate Jason-Shennen divergence from the adversarial loss) and the
VAE (with the Kulback-Leibler divergence) manifest a distribution that
is close to a normal distribution.

\hypertarget{linear-classifiers}{%
\subsection{Linear Classifiers}\label{linear-classifiers}}

Table~\ref{tbl:logreg} shows the best resulting metrics obtained with
logistic regression on the validation set, followed by the final score
on the testing set. \emph{Mix} is the identifier given for the feature
combination of SDAE and AAE. We observed that, for all classifiers, the
threshold of 0.075 would yield the best \(F_1\) score. This metric, when
obtained with the validation set, assumes the existence of only the 750
most frequent concepts in the training set. Nonetheless, these metrics
are deemed acceptable for a quantitative comparison among the trained
representations, and have indeed established the same score ordering as
the metrics in the testing set. The adversarial autoencoder obtained the
best mean \(F_1\) score in concept detection, only superseded with a
combination of the same features with those from the sparse denoising
autoencoder.

These metrics, although seemingly low, are within the expected range of
scores in the domain of concept detection in biomedical images, since
the classified labels are very scarce. As an example, only 10.9\% of the
training set is positive for the most frequent term. For the second and
third most frequent terms, the numbers are 9.8\% and 8.6\% respectively.
The mean number of positive labels of each of the 750 most frequent
concepts is 876.7, with a minimum of 203 positive labels for the 750th
most frequent concept in the training set. We find that most concepts in
the set do not have enough images with a positive label for a valuable
classifier.

The scores obtained here are on par with some of the results from the
ImageCLEF 2017 challenge. The best \(F_1\) scores on the testing set,
without the use of external resources that could severely bias the
results, were \emph{0.1583} (with a pre-trained neural network model
{[}29{]}) and \emph{0.1436} (with no external resources {[}30{]})
{[}26{]}. The use of additional information outside of the given data
sets is known to significantly improve the results. In the list of
submissions where no external resources were used, these techniques were
only outperformed by the submissions from the IPL team {[}26{]},
{[}30{]}. While the work has also relied on building global unsupervised
representations, our representations are significantly more compact in
size, and thus more computationally efficient in practice.

\hypertarget{tbl:logreg}{}
\begin{longtable}[]{@{}cccccc@{}}
\caption{\label{tbl:logreg}The best metrics obtained from logistic
regression for each representation learned, where \emph{Mix} is the
feature combination of SDAE and AAE.}\tabularnewline
\toprule
\begin{minipage}[b]{0.08\columnwidth}\centering
Type\strut
\end{minipage} & \begin{minipage}[b]{0.15\columnwidth}\centering
\(F_1\) score\strut
\end{minipage} & \begin{minipage}[b]{0.12\columnwidth}\centering
precision\strut
\end{minipage} & \begin{minipage}[b]{0.12\columnwidth}\centering
recall\strut
\end{minipage} & \begin{minipage}[b]{0.13\columnwidth}\centering
AUC\strut
\end{minipage} & \begin{minipage}[b]{0.23\columnwidth}\centering
\textbf{\(F_1\) score} (test)\strut
\end{minipage}\tabularnewline
\midrule
\endfirsthead
\toprule
\begin{minipage}[b]{0.08\columnwidth}\centering
Type\strut
\end{minipage} & \begin{minipage}[b]{0.15\columnwidth}\centering
\(F_1\) score\strut
\end{minipage} & \begin{minipage}[b]{0.12\columnwidth}\centering
precision\strut
\end{minipage} & \begin{minipage}[b]{0.12\columnwidth}\centering
recall\strut
\end{minipage} & \begin{minipage}[b]{0.13\columnwidth}\centering
AUC\strut
\end{minipage} & \begin{minipage}[b]{0.23\columnwidth}\centering
\textbf{\(F_1\) score} (test)\strut
\end{minipage}\tabularnewline
\midrule
\endhead
\begin{minipage}[t]{0.08\columnwidth}\centering
ORB\strut
\end{minipage} & \begin{minipage}[t]{0.15\columnwidth}\centering
0.13837\strut
\end{minipage} & \begin{minipage}[t]{0.12\columnwidth}\centering
0.13438\strut
\end{minipage} & \begin{minipage}[t]{0.12\columnwidth}\centering
0.14260\strut
\end{minipage} & \begin{minipage}[t]{0.13\columnwidth}\centering
0.69922\strut
\end{minipage} & \begin{minipage}[t]{0.23\columnwidth}\centering
0.09667\strut
\end{minipage}\tabularnewline
\begin{minipage}[t]{0.08\columnwidth}\centering
SIFT\strut
\end{minipage} & \begin{minipage}[t]{0.15\columnwidth}\centering
0.13303\strut
\end{minipage} & \begin{minipage}[t]{0.12\columnwidth}\centering
0.11877\strut
\end{minipage} & \begin{minipage}[t]{0.12\columnwidth}\centering
0.15118\strut
\end{minipage} & \begin{minipage}[t]{0.13\columnwidth}\centering
0.75285\strut
\end{minipage} & \begin{minipage}[t]{0.23\columnwidth}\centering
0.09518\strut
\end{minipage}\tabularnewline
\begin{minipage}[t]{0.08\columnwidth}\centering
SDAE\strut
\end{minipage} & \begin{minipage}[t]{0.15\columnwidth}\centering
0.15108\strut
\end{minipage} & \begin{minipage}[t]{0.12\columnwidth}\centering
0.14142\strut
\end{minipage} & \begin{minipage}[t]{0.12\columnwidth}\centering
0.16215\strut
\end{minipage} & \begin{minipage}[t]{0.13\columnwidth}\centering
0.78065\strut
\end{minipage} & \begin{minipage}[t]{0.23\columnwidth}\centering
0.10288\strut
\end{minipage}\tabularnewline
\begin{minipage}[t]{0.08\columnwidth}\centering
VAE\strut
\end{minipage} & \begin{minipage}[t]{0.15\columnwidth}\centering
0.13970\strut
\end{minipage} & \begin{minipage}[t]{0.12\columnwidth}\centering
0.13729\strut
\end{minipage} & \begin{minipage}[t]{0.12\columnwidth}\centering
0.14220\strut
\end{minipage} & \begin{minipage}[t]{0.13\columnwidth}\centering
0.75978\strut
\end{minipage} & \begin{minipage}[t]{0.23\columnwidth}\centering
0.09236\strut
\end{minipage}\tabularnewline
\begin{minipage}[t]{0.08\columnwidth}\centering
BiGAN\strut
\end{minipage} & \begin{minipage}[t]{0.15\columnwidth}\centering
0.14055\strut
\end{minipage} & \begin{minipage}[t]{0.12\columnwidth}\centering
0.14244\strut
\end{minipage} & \begin{minipage}[t]{0.12\columnwidth}\centering
0.13871\strut
\end{minipage} & \begin{minipage}[t]{0.13\columnwidth}\centering
0.78067\strut
\end{minipage} & \begin{minipage}[t]{0.23\columnwidth}\centering
0.09888\strut
\end{minipage}\tabularnewline
\begin{minipage}[t]{0.08\columnwidth}\centering
AAE\strut
\end{minipage} & \begin{minipage}[t]{0.15\columnwidth}\centering
\emph{0.15891}\strut
\end{minipage} & \begin{minipage}[t]{0.12\columnwidth}\centering
0.14619\strut
\end{minipage} & \begin{minipage}[t]{0.12\columnwidth}\centering
0.17406\strut
\end{minipage} & \begin{minipage}[t]{0.13\columnwidth}\centering
\emph{0.78721}\strut
\end{minipage} & \begin{minipage}[t]{0.23\columnwidth}\centering
\textbf{0.10797}\strut
\end{minipage}\tabularnewline
\begin{minipage}[t]{0.08\columnwidth}\centering
Mix\strut
\end{minipage} & \begin{minipage}[t]{0.15\columnwidth}\centering
\emph{0.16131}\strut
\end{minipage} & \begin{minipage}[t]{0.12\columnwidth}\centering
0.14704\strut
\end{minipage} & \begin{minipage}[t]{0.12\columnwidth}\centering
0.17865\strut
\end{minipage} & \begin{minipage}[t]{0.13\columnwidth}\centering
\emph{0.78865}\strut
\end{minipage} & \begin{minipage}[t]{0.23\columnwidth}\centering
\textbf{0.11050}\strut
\end{minipage}\tabularnewline
\bottomrule
\end{longtable}

The combined representation of concatenating the feature spaces of the
SDAE and AAE have resulted in even better classifiers. Although the
results of the combined representation are shown here, this improvement
is not to be overstated, given that it relies on a wider feature vector
and on training two representations that were meant to perform
individually. Another relevant observation is that the representations
based on BoWs were generally less effective for the task than deep
representation learning methods. Although SIFT BoWs have resulted in a
slightly better area under the ROC curve, the chosen operating points
led to ORB slighly outperforming SIFT.

It is understood that the thresholds can be better fine-tuned to further
increase these numbers {[}31{]}. Rather than performing a methodic
determination of the optimal threshold, we chose to avoid overfitting
the validation set by selecting a few thresholds within the interval
known to contain the optimal threshold.

\hypertarget{k-nearest-neighbors}{%
\subsection{\texorpdfstring{\(k\)-Nearest
Neighbors}{k-Nearest Neighbors}}\label{k-nearest-neighbors}}

The results of classifying the validation set with similarity search are
presented in Table~\ref{tbl:knn}. The presence of lower \(F_1\) scores
than those with linear classifiers is to be expected: the linear
classifier can be interpreted as a model which learns a custom distance
metric for each label, whereas \(k\)-NN relies on a fixed Euclidean
distance metric. With \(k\)-nearest neighbors, the best mean \(F_1\)
score of 0.07505 was obtained with the SDAE. The AAE follows with a mean
\(F_1\) score of 0.06910. The form of passive fitting over the
validation set, from the choice of \(k\), is much less greedy than the
training process of the logistic regression, which included a choice of
operating threshold and halting condition based on the outcome from the
validation set. Therefore, it is expected that the final \(F_1\) score
on the testing set heavily resembles the values obtained on the
validation set.

\hypertarget{tbl:knn}{}
\begin{longtable}[]{@{}ccccccc@{}}
\caption{\label{tbl:knn}The best \(F_1\) scores obtained from vector
similarity search for each representation learned.}\tabularnewline
\toprule
\begin{minipage}[b]{0.07\columnwidth}\centering
Type\strut
\end{minipage} & \begin{minipage}[b]{0.14\columnwidth}\centering
\(F_1\) score\strut
\end{minipage} & \begin{minipage}[b]{0.11\columnwidth}\centering
precision\strut
\end{minipage} & \begin{minipage}[b]{0.09\columnwidth}\centering
recall\strut
\end{minipage} & \begin{minipage}[b]{0.09\columnwidth}\centering
AUC\strut
\end{minipage} & \begin{minipage}[b]{0.06\columnwidth}\centering
\emph{k}\strut
\end{minipage} & \begin{minipage}[b]{0.22\columnwidth}\centering
\textbf{\(F_1\) score} (test)\strut
\end{minipage}\tabularnewline
\midrule
\endfirsthead
\toprule
\begin{minipage}[b]{0.07\columnwidth}\centering
Type\strut
\end{minipage} & \begin{minipage}[b]{0.14\columnwidth}\centering
\(F_1\) score\strut
\end{minipage} & \begin{minipage}[b]{0.11\columnwidth}\centering
precision\strut
\end{minipage} & \begin{minipage}[b]{0.09\columnwidth}\centering
recall\strut
\end{minipage} & \begin{minipage}[b]{0.09\columnwidth}\centering
AUC\strut
\end{minipage} & \begin{minipage}[b]{0.06\columnwidth}\centering
\emph{k}\strut
\end{minipage} & \begin{minipage}[b]{0.22\columnwidth}\centering
\textbf{\(F_1\) score} (test)\strut
\end{minipage}\tabularnewline
\midrule
\endhead
\begin{minipage}[t]{0.07\columnwidth}\centering
ORB\strut
\end{minipage} & \begin{minipage}[t]{0.14\columnwidth}\centering
0.04279\strut
\end{minipage} & \begin{minipage}[t]{0.11\columnwidth}\centering
0.02990\strut
\end{minipage} & \begin{minipage}[t]{0.09\columnwidth}\centering
0.10575\strut
\end{minipage} & \begin{minipage}[t]{0.09\columnwidth}\centering
0.55240\strut
\end{minipage} & \begin{minipage}[t]{0.06\columnwidth}\centering
4\strut
\end{minipage} & \begin{minipage}[t]{0.22\columnwidth}\centering
0.04177\strut
\end{minipage}\tabularnewline
\begin{minipage}[t]{0.07\columnwidth}\centering
SIFT\strut
\end{minipage} & \begin{minipage}[t]{0.14\columnwidth}\centering
0.05980\strut
\end{minipage} & \begin{minipage}[t]{0.11\columnwidth}\centering
0.04341\strut
\end{minipage} & \begin{minipage}[t]{0.09\columnwidth}\centering
0.13402\strut
\end{minipage} & \begin{minipage}[t]{0.09\columnwidth}\centering
0.56660\strut
\end{minipage} & \begin{minipage}[t]{0.06\columnwidth}\centering
3\strut
\end{minipage} & \begin{minipage}[t]{0.22\columnwidth}\centering
0.05665\strut
\end{minipage}\tabularnewline
\begin{minipage}[t]{0.07\columnwidth}\centering
SDAE\strut
\end{minipage} & \begin{minipage}[t]{0.14\columnwidth}\centering
\emph{0.07995}\strut
\end{minipage} & \begin{minipage}[t]{0.11\columnwidth}\centering
0.06954\strut
\end{minipage} & \begin{minipage}[t]{0.09\columnwidth}\centering
0.12041\strut
\end{minipage} & \begin{minipage}[t]{0.09\columnwidth}\centering
0.55998\strut
\end{minipage} & \begin{minipage}[t]{0.06\columnwidth}\centering
2\strut
\end{minipage} & \begin{minipage}[t]{0.22\columnwidth}\centering
\textbf{0.07505}\strut
\end{minipage}\tabularnewline
\begin{minipage}[t]{0.07\columnwidth}\centering
VAE\strut
\end{minipage} & \begin{minipage}[t]{0.14\columnwidth}\centering
0.03569\strut
\end{minipage} & \begin{minipage}[t]{0.11\columnwidth}\centering
0.02490\strut
\end{minipage} & \begin{minipage}[t]{0.09\columnwidth}\centering
0.08703\strut
\end{minipage} & \begin{minipage}[t]{0.09\columnwidth}\centering
0.54307\strut
\end{minipage} & \begin{minipage}[t]{0.06\columnwidth}\centering
4\strut
\end{minipage} & \begin{minipage}[t]{0.22\columnwidth}\centering
0.03450\strut
\end{minipage}\tabularnewline
\begin{minipage}[t]{0.07\columnwidth}\centering
BiGAN\strut
\end{minipage} & \begin{minipage}[t]{0.14\columnwidth}\centering
0.04700\strut
\end{minipage} & \begin{minipage}[t]{0.11\columnwidth}\centering
0.03496\strut
\end{minipage} & \begin{minipage}[t]{0.09\columnwidth}\centering
0.09928\strut
\end{minipage} & \begin{minipage}[t]{0.09\columnwidth}\centering
0.54926\strut
\end{minipage} & \begin{minipage}[t]{0.06\columnwidth}\centering
3\strut
\end{minipage} & \begin{minipage}[t]{0.22\columnwidth}\centering
0.04733\strut
\end{minipage}\tabularnewline
\begin{minipage}[t]{0.07\columnwidth}\centering
AAE\strut
\end{minipage} & \begin{minipage}[t]{0.14\columnwidth}\centering
0.07212\strut
\end{minipage} & \begin{minipage}[t]{0.11\columnwidth}\centering
0.06263\strut
\end{minipage} & \begin{minipage}[t]{0.09\columnwidth}\centering
0.10898\strut
\end{minipage} & \begin{minipage}[t]{0.09\columnwidth}\centering
0.55427\strut
\end{minipage} & \begin{minipage}[t]{0.06\columnwidth}\centering
2\strut
\end{minipage} & \begin{minipage}[t]{0.22\columnwidth}\centering
0.06910\strut
\end{minipage}\tabularnewline
\bottomrule
\end{longtable}

\hypertarget{conclusion}{%
\section{Conclusion}\label{conclusion}}

This paper takes unsupervised representation learning techniques from
state-of-the-art, facing them against a more traditional bags of visual
words approach. The methods were evaluated with the biomedical concept
detection sub-task of the ImageCLEF 2017 caption prediction task. We
have tested the hypothesis that a powerful image descriptor can
contribute to efficient concept detection with some level of certainty,
without observing the original image. Results are presented for six
different approaches, where two of them rely on visual keypoint
extraction and description algorithms, and other two of them are based
on generative adversarial networks. Overall, these methods have
significantly outperformed our previous participation and are on par
with other techniques in the challenge.

As identified in {[}32{]} and proved in this work, it is possible to
obtain more powerful representations with modern deep learning
approaches, in contrast with previously popular computer vision methods
such as the SIFT bags of visual words. Deep learning techniques based on
GANs can provide good results, but the additional complexity, the
difficulty of convergence, and the possibility of mode collapse can
significantly cripple their performance in representation learning.
Nonetheless, these issues are already a high focus of attentions at this
time, and will likely lead to substantial improvements in GAN design and
training.

It is also important that these approaches are augmented with non-visual
information. In particular, a medical imaging archive should take the
most advantage of the available data beyond pixel data. Future work will
consider semi-supervised learning as a means of building more
descriptive representations from known categories and other annotations.
Subsequently, these representations are to be evaluated in a medical
information retrieval scenario, as well as with other data sets in the
medical imaging domain.

\hypertarget{acknowledgements}{%
\section{Acknowledgements}\label{acknowledgements}}

This work is financed by the ERDF - European Regional Development Fund
through the Operational Programme for Competitiveness and
Internationalisation - COMPETE 2020 Programme, and by National Funds
through the FCT -- Fundação para a Ciência e a Tecnologia, within
project PTDC/EEI-ESS/6815/2014. Eduardo Pinho is funded by the FCT under
the grant PD/BD/105806/2014.

\hypertarget{references}{%
\section*{References}\label{references}}
\addcontentsline{toc}{section}{References}

\hypertarget{refs}{}
\leavevmode\hypertarget{ref-bengio2013representation}{}%
{[}1{]} Y. Bengio, A. Courville, and P. Vincent, ``Representation
learning: A review and new perspectives,'' \emph{Pattern Analysis and
Machine Intelligence, IEEE Transactions on}, vol. 35, no. 8, pp.
1798--1828, 2013.

\leavevmode\hypertarget{ref-ravi2017deep}{}%
{[}2{]} D. Ravi \emph{et al.}, ``Deep Learning for Health Informatics,''
\emph{Biomedical and Health Informatics, IEEE Journal of}, vol. 21, no.
1, pp. 4--21, Jan. 2017.

\leavevmode\hypertarget{ref-lee2006efficient}{}%
{[}3{]} H. Lee, A. Battle, R. Raina, and A. Y. Ng, ``Efficient sparse
coding algorithms,'' \emph{Advances in Neural Information Processing
Systems}, vol. 19, no. 2, pp. 801--808, 2006.

\leavevmode\hypertarget{ref-hinton2006fast}{}%
{[}4{]} G. E. Hinton, S. Osindero, and Y.-W. Teh, ``A fast learning
algorithm for deep belief nets,'' \emph{Neural computation}, vol. 18,
no. 7, pp. 1527--1554, 2006.

\leavevmode\hypertarget{ref-wangming2008application}{}%
{[}5{]} X. Wangming, W. Jin, L. Xinhai, Z. Lei, and S. Gang,
``Application of Image SIFT Features to the Context of CBIR,'' in
\emph{International conference on computer science and software
engineering}, 2008, pp. 552--555.

\leavevmode\hypertarget{ref-dimitrovski2015improved}{}%
{[}6{]} I. Dimitrovski, D. Kocev, I. Kitanovski, S. Loskovska, and S.
Džeroski, ``Improved medical image modality classification using a
combination of visual and textual features,'' \emph{Computerized Medical
Imaging and Graphics}, vol. 39, pp. 14--26, 2015.

\leavevmode\hypertarget{ref-vincent2010stacked}{}%
{[}7{]} P. Vincent, H. Larochelle, I. Lajoie, Y. Bengio, and P.-a.
Manzagol, ``Stacked Denoising Autoencoders: Learning Useful
Representations in a Deep Network with a Local Denoising Criterion,''
\emph{Journal of Machine Learning Research}, vol. 11, pp. 3371--3408,
2010.

\leavevmode\hypertarget{ref-kingma2014auto}{}%
{[}8{]} D. P. Kingma and M. Welling, ``Auto-Encoding Variational
Bayes,'' pp. 1--14, 2014.

\leavevmode\hypertarget{ref-goodfellow2014generative}{}%
{[}9{]} I. J. Goodfellow \emph{et al.}, ``Generative Adversarial Nets,''
pp. 1--9, 2014.

\leavevmode\hypertarget{ref-donahue2016adversarial}{}%
{[}10{]} J. Donahue, P. Krähenbühl, and T. Darrell, ``Adversarial
feature learning,'' \emph{arXiv preprint arXiv:1605.09782}, May 2016.

\leavevmode\hypertarget{ref-li2017large}{}%
{[}11{]} Z. Li, X. Zhang, H. Müller, and S. Zhang, ``Large-scale
Retrieval for Medical Image Analytics: A Comprehensive Review,''
\emph{Medical Image Analysis}, 2017.

\leavevmode\hypertarget{ref-kalpathy2015evaluating}{}%
{[}12{]} J. Kalpathy-Cramer, A. G. S. de Herrera, D. Demner-Fushman, S.
Antani, S. Bedrick, and H. Müller, ``Evaluating performance of
biomedical image retrieval systems---an overview of the medical image
retrieval task at ImageCLEF 2004--2013,'' \emph{Computerized Medical
Imaging and Graphics}, vol. 39, pp. 55--61, 2015.

\leavevmode\hypertarget{ref-lijens2017survey}{}%
{[}13{]} G. Litjens \emph{et al.}, ``A survey on deep learning in
medical image analysis,'' \emph{Medical Image Analysis}, vol. 42, pp.
60--88, 2017.

\leavevmode\hypertarget{ref-sun2017automatic}{}%
{[}14{]} W. Sun, B. Zheng, and W. Qian, ``Automatic feature learning
using multichannel ROI based on deep structured algorithms for
computerized lung cancer diagnosis,'' \emph{Computers in biology and
medicine}, vol. 89, pp. 530--539, 2017.

\leavevmode\hypertarget{ref-wu2016scalable}{}%
{[}15{]} G. Wu, M. Kim, Q. Wang, B. C. Munsell, and D. Shen, ``Scalable
high-performance image registration framework by unsupervised deep
feature representations learning,'' \emph{IEEE Transactions on
Biomedical Engineering}, vol. 63, no. 7, pp. 1505--1516, 2016.

\leavevmode\hypertarget{ref-lowe2004distinctive}{}%
{[}16{]} D. G. Lowe, ``Distinctive Image Features from Scale-Invariant
Keypoints,'' \emph{International Journal of Computer Vision}, vol. 60,
no. 2, pp. 91--110, 2004.

\leavevmode\hypertarget{ref-rublee2011orb}{}%
{[}17{]} E. Rublee, V. Rabaud, K. Konolige, and G. Bradski, ``ORB: An
efficient alternative to SIFT or SURF,'' in \emph{Computer vision
(ICCV), 2011 IEEE international conference on}, 2011, pp. 2564--2571.

\leavevmode\hypertarget{ref-bradski2000opencv}{}%
{[}18{]} G. Bradski and Others, ``The OpenCV library,'' \emph{Doctor
Dobbs Journal}, vol. 25, no. 11, pp. 120--126, 2000.

\leavevmode\hypertarget{ref-sivic2003video}{}%
{[}19{]} J. Sivic and A. Zisserman, ``Video google: A text retrieval
approach to object matching in videos,'' in \emph{Computer vision, 2003.
Proceedings. Ninth IEEE international conference on}, 2003, pp.
1470--1477.

\leavevmode\hypertarget{ref-radford2016unsupervised}{}%
{[}20{]} A. Radford, L. Metz, and S. Chintala, ``Unsupervised
Representation Learning with Deep Convolutional Generative Adversarial
Networks,'' \emph{arXiv preprint arXiv:1511.06434}, pp. 1--16, 2016.

\leavevmode\hypertarget{ref-ioffe2015batch}{}%
{[}21{]} S. Ioffe and C. Szegedy, ``Batch Normalization: Accelerating
Deep Network Training by Reducing Internal Covariate Shift,''
\emph{arXiv preprint arXiv:1502.03167}, pp. 1--11, 2015.

\leavevmode\hypertarget{ref-ba2016layer}{}%
{[}22{]} J. L. Ba, J. R. Kiros, and G. E. Hinton, ``Layer
normalization,'' Jul. 2016.

\leavevmode\hypertarget{ref-makhzani2016adversarial}{}%
{[}23{]} A. Makhzani, J. Shlens, N. Jaitly, I. Goodfellow, and B. Frey,
``Adversarial Autoencoders,'' Nov. 2015.

\leavevmode\hypertarget{ref-kingma2015adam}{}%
{[}24{]} D. P. Kingma and J. L. Ba, ``Adam: A Method for Stochastic
Optimization,'' in \emph{International conference on learning
representations}, 2015.

\leavevmode\hypertarget{ref-abadi2016tensorflow}{}%
{[}25{]} M. Abadi \emph{et al.}, ``TensorFlow: Large-Scale Machine
Learning on Heterogeneous Distributed Systems,'' Mar. 2016.

\leavevmode\hypertarget{ref-imageclefoverview2017caption}{}%
{[}26{]} C. Eickhoff, I. Schwall, A. de Herrera, and H. Müller,
``Overview of ImageCLEFcaption 2017 - the image caption prediction and
concept extraction tasks to understand biomedical images,'' \emph{CLEF
working notes, CEUR}, 2017.

\leavevmode\hypertarget{ref-mcmahan2013ad}{}%
{[}27{]} H. B. McMahan \emph{et al.}, ``Ad click prediction: a view from
the trenches,'' in \emph{Proceedings of the 19th acm sigkdd
international conference on knowledge discovery and data mining}, 2013,
pp. 1222--1230.

\leavevmode\hypertarget{ref-johnson2017billion}{}%
{[}28{]} J. Johnson, M. Douze, and H. Jégou, ``Billion-scale similarity
search with GPUs,'' \emph{arXiv preprint arXiv:1702.08734}, 2017.

\leavevmode\hypertarget{ref-dimitris2017concept}{}%
{[}29{]} K. Dimitris and K. Ergina, ``Concept detection on medical
images using Deep Residual Learning Network,'' in \emph{Working notes of
conference and labs of the evaluation forum}, 2017.

\leavevmode\hypertarget{ref-valavanis2017ipl}{}%
{[}30{]} L. Valavanis and S. Stathopoulos, ``IPL at ImageCLEF 2017
Concept Detection Task,'' in \emph{Working notes of conference and labs
of the evaluation forum}, 2017.

\leavevmode\hypertarget{ref-lipton2014thresholding}{}%
{[}31{]} Z. C. Lipton, C. Elkan, and B. Narayanaswamy, ``Thresholding
Classifiers to Maximize F1 Score,'' \emph{Machine Learning and Knowledge
Discovery in Databases}, vol. 8725, pp. 225-----239, Feb. 2014.

\leavevmode\hypertarget{ref-pinho2017representation}{}%
{[}32{]} E. Pinho, J. F. Silva, J. M. Silva, and C. Costa, ``Towards
Representation Learning for Biomedical Concept Detection in Medical
Images: UA. PT Bioinformatics in ImageCLEF 2017,'' in \emph{Working
notes of conference and labs of the evaluation forum}, 2017.

\end{document}